\documentclass[journal]{IEEEtran}
\usepackage{amsmath,amsfonts}
\usepackage{algorithm}
\usepackage{array}
\usepackage[caption=false,font=normalsize,labelfont=sf,textfont=sf]{subfig}
\usepackage{textcomp}
\usepackage{stfloats}
\usepackage{url}
\usepackage{verbatim}
\usepackage{cite}
\usepackage[colorlinks,linkcolor=red,citecolor=blue]{hyperref}

\ifCLASSINFOpdf
\usepackage[pdftex]{graphicx}
\DeclareGraphicsExtensions{.eps,.pdf,.tiff,.jpeg,.png}
\else
\usepackage[dvips]{graphicx}
\DeclareGraphicsExtensions{.eps}
\fi
\graphicspath{{pic/}{pic/authors}{Major/pic}{Major/pic/authors}}

\usepackage{graphicx, algcompatible, epstopdf,epsfig,amssymb,multirow, enumerate,float, diagbox,makecell}
\usepackage[caption=false]{subfig}

\begin{document}
	
	\title{Text-Driven Traffic Anomaly Detection with Temporal High-Frequency Modeling in Driving Videos}
	
	\author{{Rongqin Liang,~\IEEEmembership{Student Member,~IEEE,} 
			Yuanman Li,~\IEEEmembership{Senior Member,~IEEE,} 
			Jiantao Zhou,~\IEEEmembership{Senior Member,~IEEE,} and 
			Xia Li,~\IEEEmembership{Member,~IEEE}}
		\thanks{This work was supported in part by in part by the Key project of Shenzhen Science and Technology Plan under Grant 20220810180617001 and the Foundation for Science and Technology Innovation of Shenzhen under Grant RCBS20210609103708014; in part by the Guangdong Basic and Applied Basic Research Foundation under Grant 2022A1515010645;  in part by the Open Research Project Programme of the State Key Laboratory of Internet of Things for Smart City (University of Macau) under Grant SKLIoTSC(UM)-2021-2023/ORP/GA04/2022. (Corresponding author: Yuanman Li)
			
			Rongqin Liang, Yuanman Li and Xia Li are with Guangdong Key Laboratory of Intelligent Information Processing, College of Electronics and Information Engineering, Shenzhen University, Shenzhen 518060, China (email: 1810262064@email.szu.edu.cn; yuanmanli@szu.edu.cn; lixia@szu.edu.cn).
			
			Jiantao Zhou is with the State Key Laboratory of Internet of Things for Smart City and the Department of Computer and Information Science, University of Macau, Macau (e-mail: jtzhou@um.edu.mo).}
	}
	
	\markboth{IEEE TRANSACTIONS ON CIRCUITS AND SYSTEMS FOR VIDEO TECHNOLOGY}%
	{Shell \MakeLowercase{\textit{et al.}}: A Sample Article Using IEEEtran.cls for IEEE Journals}
    \IEEEpubid{\begin{minipage}{\textwidth}\ \centering
    Copyright © 2024 IEEE. Personal use of this material is permitted. \\
    However, permission to use this material for any other purposes must be obtained from the IEEE by sending an email to pubs-permissions@ieee.org.
    \end{minipage}}
 
	\maketitle
	
	\begin{abstract}
		Traffic anomaly detection (TAD) in driving videos is critical for ensuring the safety of autonomous driving and advanced driver assistance systems.  Previous single-stage TAD methods primarily rely on frame prediction, making them vulnerable to interference from dynamic backgrounds induced by the rapid movement of the dashboard camera. While two-stage TAD methods appear to be a natural solution to mitigate such interference by pre-extracting background-independent features (such as bounding boxes and optical flow) using perceptual algorithms, they are susceptible to the performance of first-stage perceptual algorithms and may result in error propagation. 
		In this paper, we introduce TTHF, a novel single-stage method aligning video clips with text prompts, offering a new perspective on traffic anomaly detection. Unlike previous approaches, the supervised signal of our method is derived from languages rather than orthogonal one-hot vectors, providing a more comprehensive representation. Further, concerning visual representation, we propose to model the high frequency of driving videos in the temporal domain. This modeling captures the dynamic changes of driving scenes, enhances the perception of driving behavior, and significantly improves the detection of traffic anomalies. 
		In addition, to better perceive various types of traffic anomalies, we carefully design an attentive anomaly focusing mechanism that visually and linguistically guides the model to adaptively focus on the visual context of interest, thereby facilitating the detection of traffic anomalies.
		It is shown that our proposed TTHF achieves promising performance, outperforming state-of-the-art competitors by +5.4\% AUC on the DoTA dataset and achieving high generalization on the DADA dataset.
	\end{abstract}
	
	\begin{IEEEkeywords}
		Traffic anomaly detection, multi-modality learning, high frequency, attention.
	\end{IEEEkeywords}
	
	\section{Introduction}
	\IEEEPARstart{T}{RAFFIC} anomaly detection (TAD) in driving videos is a crucial component of automated driving systems \cite{9438625, 8715479} and advanced driver assistance systems \cite{7428841, 8809914}. It is designed to detect anomalous traffic behavior from the first-person driving perspective. Accurate detection of traffic anomalies helps improve road safety, shorten traffic recovery times, and reduce the number of regrettable daily traffic accidents.
	\begin{figure}[!t]
		\centering
		\includegraphics[width=0.90\linewidth]{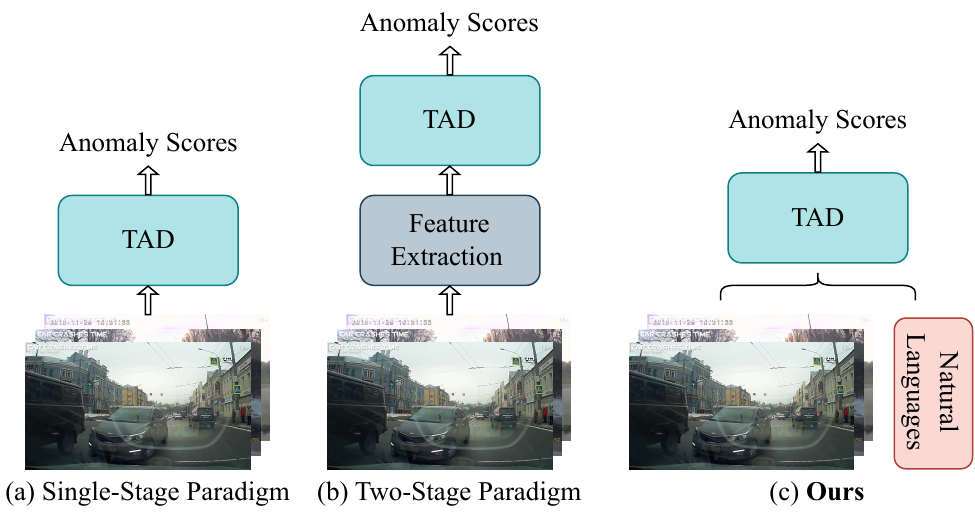}
		\caption{Existing TAD approaches of single-stage paradigm (a) and two-stage paradigm (b) vs. the proposed TTHF framework (c). Existing single-stage approaches mainly rely on frame prediction, which is difficult to adapt to detecting traffic anomalies with a dynamic background, while the two-stage TAD approaches are vulnerable to the performance of the first-stage perceptual algorithms. The proposed TTHF framework is text-driven and focuses on capturing dynamic changes in driving scenes through modeling temporal high frequency to facilitate traffic anomaly detection.}
		\label{fig:paradigm}
	\end{figure}
	\IEEEpubidadjcol
 
	Given the significance of traffic anomaly detection, scholars are actively involved in this field and have proposed constructive research \cite{7564410, Liu_2018_CVPR, Liu_2021_ICCV, 9714213, 9712446}. We observe that these works on TAD can be mainly divided into the single-stage paradigm \cite{Hasan_2016_CVPR, 10.1007, Liu_2018_CVPR} and the two-stage paradigm \cite{9714213, 9712446, 9733965}. As shown in Fig. \ref{fig:paradigm}, previous TAD methods mainly embrace a single-stage paradigm, exemplified by frame prediction \cite{Liu_2018_CVPR} and reconstruction-based \cite{10.1007} TAD approaches. Nevertheless, these methods are subject to the dynamic backgrounds caused by the rapid movement of the dashboard camera and have limited accuracy in detecting traffic anomalies. To confront the challenges posed by dynamic backgrounds, researchers have advocated for TAD methods \cite{9714213, 9712446, 9733965} that utilize a two-stage paradigm. These two-stage approaches first extract features such as optical flow, bounding boxes, or tracking IDs from video frames using existing visual perception algorithms, and then propose a TAD model for detecting traffic anomalies. While these approaches have laid the foundation for TAD in driving videos, they are susceptible to the performance of the first-stage visual perception algorithm, which may cause error propagation, resulting in false detection or missing traffic anomalies. Therefore, in this paper, we strive to explore an effective single-stage paradigm-based approach for traffic anomaly detection in driving videos. 
	
	Recently, large-scale visual language pre-training models \cite{pmlr-v139-radford21a, pmlr-v139-jia21b, Yang_2023_ICCV} have achieved remarkable results by utilizing language knowledge to assist with visual tasks. Among them, CLIP \cite{pmlr-v139-radford21a} stands out for its exceptional transferability through the alignment of image-text semantics and has demonstrated outstanding capabilities across various computer vision tasks such as object detection \cite{gu2022openvocabulary}, semantic segmentation \cite{Xu_2022_CVPR}, and video retrieval \cite{10268993}. The success of image-text alignment techniques can be attributed to their ability to map the natural languages associated with an image into high-dimensional non-orthogonal vectors. This is in contrast to traditional supervised methods that map predefined labels to low-dimensional one-hot vectors. Compared to the low-dimensional one-hot vectors, these high-dimensional vectors offer more comprehensive representations to guide the network training.
	Motivated by this, we endeavor to investigate a language-guided approach for detecting traffic anomalies in driving videos. Intuitively, the transition of CLIP from image-text alignment to video-text alignment primarily involves the consideration of modeling temporal dimensions. Despite the exploration of various methods \cite{wang2021actionclip, LUO2022293, Rasheed_2023_CVPR, 10413510} for temporal modeling, encompassing various techniques such as \textit{Average Pooling}, \textit{Conv1D}, \textit{LSTM}, \textit{Transformer}, the existing approaches predominantly concentrate on aggregating visual context along the temporal dimension. In the context of traffic anomaly detection for driving videos, we emphasize that beyond the visual context, characterizing dynamic changes in the driving scene along the temporal dimension proves advantageous in determining abnormal driving behavior. For instance, traffic events such as vehicle collisions or loss of control often result in significant and rapid alterations in the driving scene. Therefore, \textit{how to effectively characterize the dynamic changes of driving scenes holds paramount importance for traffic anomaly detection in driving videos}.
	
	Additionally, considering that different types of traffic anomalies exhibit unique characteristics, a straightforward encoding of the entire driving scene may diminish the discriminability of driving events and impede the detection of diverse traffic anomalies. For instance, traffic anomalies involving the ego-vehicle are often accompanied by global jittering of the dashboard camera, while anomalies involving non-ego vehicles often lead to local anomalies in the driving scene. Consequently, \textit{how to better perceive various types of traffic anomalies proves crucial for traffic anomaly detection}.
	
	In this work, we propose a novel traffic anomaly detection approach: \textbf{T}ext-Driven Traffic Anomaly Detection with \textbf{T}emporal \textbf{H}igh-\textbf{F}requency Modeling (TTHF), as shown in Fig. \ref{fig:framework}. 
	To represent driving videos comprehensively, our fundamental idea is to not only capture the spatial visual context but also emphasize the depiction of dynamic changes in the driving scenes, thereby enhancing the visual representation of driving videos. Specifically, we initially leverage the pre-trained visual encoder of CLIP, endowed with rich prior knowledge of visual language semantics, to encode the visual context of driving videos. Then, to capture the dynamic changes in driving scenes, we innovatively introduce temporal high-frequency modeling (THFM) to obtain temporal high frequency representations of driving videos along the temporal dimension. Subsequently, the visual context and temporal high-frequency representations are fused to enhance the overall visual representation of driving videos. 
	To better perceive various types of traffic anomalies, we propose an attentive anomaly focusing mechanism (AAFM) to guide the model to adaptively focus both visually and linguistically on the visual context of interest, thereby facilitating the detection of traffic anomalies.
	
	It is shown that our proposed TTHF model exhibits promising performance on the DoTA dataset \cite{9712446}, outperforming state-of-the-art competitors by +5.4\% AUC. Furthermore, without any fine tuning, the AUC performance of TTHF on the DADA dataset \cite{9312486} demonstrates its generalization capability. The main contributions of our work can be summarized as follows:
	\begin{enumerate}
		\item We introduce a simple yet effective single-stage traffic anomaly detection method that aligns the visual semantics of driving videos with matched textual semantics to identify traffic anomalies. In contrast to previous TAD methods, the supervised signals in our approach are derived from text, offering a more comprehensive representation in high-dimensional space.
		\item We emphasize the modeling of high frequency in the temporal domain for driving videos. In contrast to previous approaches that solely aggregate visual context along the temporal dimension, we place additional emphasis on modeling high frequency in the temporal domain. This enables us to characterize dynamic changes in the driving scene over time, thereby significantly enhancing the performance of traffic anomaly detection.
		\item We further propose an attentive anomaly focusing mechanism to enhance the perception of various traffic anomalies. Our proposed mechanism guides the model both visually and linguistically to adaptively focus on the visual contexts of interest, facilitating the detection of traffic anomalies.
		\item Comprehensive experimental results on public benchmark datasets demonstrate the superiority and robustness of the proposed method. Compared to existing state-of-the-art methods, the proposed TTHF improves AUC by +5.4\% on the DoTA dataset and also achieves state-of-the-art AUC on the DADA dataset without any fine-tuning.
	\end{enumerate}
	
	The remainder of this paper is organized as follows. Section \ref{related_work} gives a brief review of related works. Section \ref{sec:proposed} details our proposed TTHF for traffic anomaly detection in driving videos. Extensive experimental results are presented in Section \ref{sec:experiment}, and we finally draw a conclusion in Section \ref{sec:conclusion}.
	\section{Related Works} \label{related_work}
	\subsection{Traffic Anomaly Detection (TAD) in Driving Videos}
	Traffic anomaly detection (TAD) in driving videos aims to identify abnormal traffic events from the perspective of driving, such as collisions with other vehicles or obstacles, being out of control, and so on. Such events can be classified into two categories: ego-involved anomalies (\textit{i.e.}, traffic events involving the ego-vehicle) and non-ego anomalies (\textit{i.e.}, traffic events involving observed objects but not the ego-vehicle). A closely related topic to TAD in driving videos is anomaly detection in surveillance videos (VAD), which involves identifying abnormal events such as fights, assaults, thefts, arson, and so forth from a surveillance viewpoint. In recent years, various VAD methods \cite{9828496, 9410375, 9701300, 9645572, 9749781, 9860012} have been proposed for surveillance videos, which have greatly contributed to the development of this field. However, in contrast to the static background in surveillance videos, the background in driving videos is dynamically changing due to the fast movement of the ego vehicle, which makes the VAD methods prone to failure in the TAD task \cite{9712446, 9733965}. Recently, Wang et al. \cite{10417748} proposed a method for detecting crowd flow anomalies by comparing anomalous samples with normal samples that were estimated based on prototypes. However, crowd flow anomaly detection methods are difficult to apply to the TAD task due to the differences in tasks and the data processed. In this paper, we work on the task of traffic anomaly detection in driving videos to provide a new solution for this community.
	
	Early TAD methods \cite{7564410, YUAN2018202} mainly extracted features in a handcrafted manner and utilized a Bayesian model for classification. However, these methods are sensitive to well-designed features and generally lack robustness in dealing with a wide variety of traffic scenarios. With the advances of deep neural networks in computer vision, researchers have proposed deep learning-based approaches for TAD, laying the foundation for this task. Based on our observations, the existing TAD methods can be basically classified into single-stage paradigm \cite{Hasan_2016_CVPR, 10.1007, Liu_2018_CVPR} and two-stage paradigm \cite{8967556, sun2022anomaly, 9733965, liang2023memoryaugmented}. 
	
	Previous single-stage paradigm-based TAD approaches mainly comprise frame reconstruction-based and frame prediction-based TAD approaches \cite{Hasan_2016_CVPR, 10.1007, Liu_2018_CVPR}. These methods used reconstruction or prediction errors of video frames to evaluate traffic anomalies. For instance, Liu et al. \cite{Liu_2018_CVPR} predicted video frames of normal traffic events through appearance and motion constraints, thereby helping to identify traffic anomalies that do not conform to expectations. Unfortunately, these methods tend to detect ego-involved anomalies (\textit{e.g.}, out of control) and perform poorly on non-ego traffic anomalies. This is primarily attributed to ego-involved anomalies causing significant shaking of the dashboard camera, leading to substantial global errors in frame reconstruction or prediction. Such errors undoubtedly facilitate anomaly detection. However, the methods based on frame reconstruction or prediction have difficulty distinguishing the local errors caused by the traffic anomalies of other road participants because of the interference of the dynamic background from the fast-moving ego-vehicle. This impairs their ability to detect traffic anomalies.
	
	In recent years, to address the challenges posed by dynamic backgrounds, researchers have proposed applying a two-stage paradigm to the traffic anomaly detection task. In this paradigm, the perception algorithm is initially applied to extract visual features in the first stage. Then, the TAD model utilizes these features to detect traffic anomalies. For instance, Yao et al. \cite{9712446, 8967556} applied Mask-RCNN \cite{He_2017_ICCV}, FlowNet \cite{Ilg_2017_CVPR}, DeepSort \cite{8296962}, and ORBSLAM \cite{7946260} algorithms to extract bounding boxes (bboxes), optical flow, tracking ids, and ego motion, respectively. Then, they used these visual features to predict the future locations of objects over a short horizon and detected traffic anomalies based on the deviation of the predicted location. Along this line, Fang et al. \cite{9733965} used optical flow and bboxes as visual features. They attempted to collaborate on frame prediction and future object localization tasks \cite{10195882} to detect traffic anomalies by analyzing inconsistencies in predicted frames, object locations, and the spatial relation structure of the scene. Zhou et al. \cite{9714213} obtained bboxes of objects in the scene from potentially abnormal frames as visual features. They then encoded the spatial relationships of the detected objects to determine the abnormality of these frames. Despite the success of the two-stage paradigm TAD methods, they rely on the perception algorithms in the first stage, which may cause error propagation and lead to missed or false detection of traffic anomalies. Different from existing TAD methods, we propose a text-driven single-stage traffic anomaly detection approach that provides a promising solution for this task.
	\begin{figure*}[!t]
		\centering
		\includegraphics[width=0.88\linewidth]{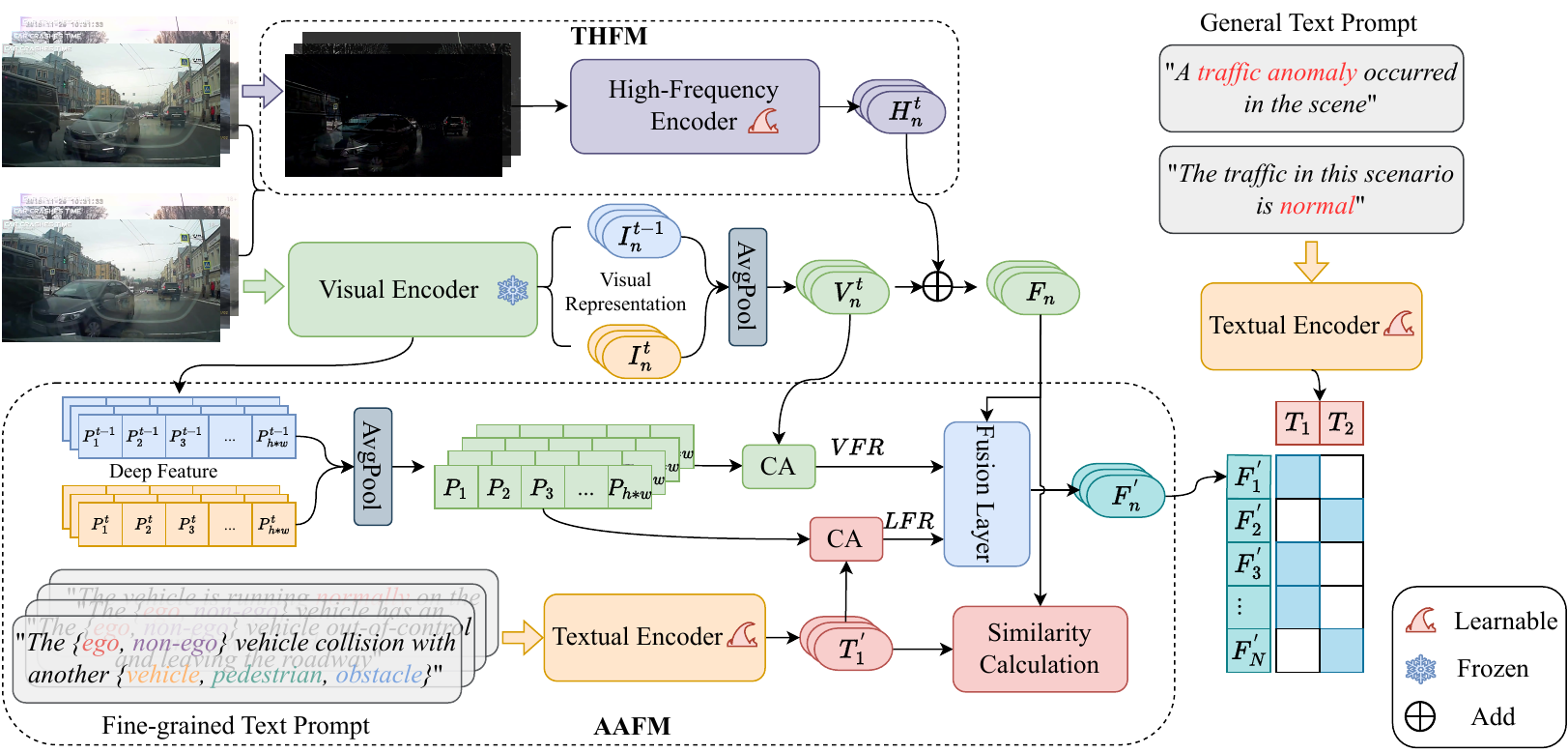}
		\caption{Overview of our proposed TTHF. It is a CLIP-like framework for traffic anomaly detection. In this framework, we first apply a visual encoder to extract visual representations of driving video clips. Then, we propose Temporal High-Frequency Modeling (THFM) to characterize the dynamic changes of driving scenes and thus construct a more comprehensive representation of driving videos. Finally, we introduce an attentive anomaly focusing mechanism (AAFM) to enhance the perception of various types of traffic anomalies. Besides, for brevity, we denote the cross-attention as CA, the visually focused representation as VFR, and the linguistically focused representation as LFR.}
		\label{fig:framework}
	\end{figure*}
	\subsection{Vision-Text Multi-Modality Learning}
	Recently, there has been a gradual focus on vision-text multi-modal learning. Among them, contrastive language-image pre-training methods have achieved remarkable results in many computer vision tasks such as image classification \cite{pmlr-v139-radford21a, pmlr-v139-jia21b}, object detection \cite{gu2022openvocabulary, Yao_2023_CVPR}, semantic segmentation \cite{Xu_2022_CVPR, Zhou_2023_CVPR} and image retrieval \cite{Baldrati_2023_ICCV, Tschannen_2023_CVPR}. At present, CLIP \cite{pmlr-v139-radford21a} has become a mainstream visual learning method, which connects visual signals and language semantics by comparing large-scale image-language pairs. Essentially, compared to traditional supervised methods that convert labels into orthogonal one-hot vectors, CLIP provides richer and more comprehensive supervision information by collecting large-scale image-text pairs from web data and mapping the text into high-dimensional supervision signals (usually non-orthogonal). Following this idea, many scholars have applied CLIP to various tasks in the video domain, including video action recognition \cite{wang2021actionclip, 10.1007/978-3-031-20062-5_39}, video retrieval \cite{10268993, LUO2022293, 10.1145/3503161.3547910}, video recognition \cite{Wu_Sun_Ouyang_2023, 10.1007/978-3-031-19772-7_1}, and so on. For example, ActionCLIP \cite{wang2021actionclip} modeled the video action detection task as a video-text matching problem in a multi-modal learning framework and strengthened the video representation with more semantic language supervision to enable the model to perform zero-shot action recognition. 
	More recently, Wu et al. \cite{wu2023vadclip} proposed a vision-language model for anomaly detection in surveillance videos. However, as mentioned earlier, traffic anomaly detection faces the problem of dynamic changes in the driving scene, which often makes VAD methods prone to fail in TAD tasks. To the best of our knowledge, there is no effective approach to model traffic anomaly detection task from the perspective of vision-text multi-modal learning. In this paper, we preliminarily explore an effective text-driven method for traffic anomaly detection, which we hope can provide a new perspective on this task.
	\section{The Proposed Approach: TTHF} \label{sec:proposed}
	In this section, we mainly introduce the proposed TTHF framework. First, we describe the overall framework of TTHF. Then, we explain two key modules in TTHF, \textit{i.e.}, temporal High-Frequency Modeling (THFM) and attentive anomaly focusing mechanism (AAFM). Moreover, we describe the contrastive learning strategy for cross-modal learning of video-text pairs, and finally show how to perform traffic anomaly detection in our TTHF.
	
	\subsection{Overview of Our TTHF Framework}
	The overall framework of TTHF is illustrated in Fig. \ref{fig:framework}. It presents a CLIP-like two-stream framework for traffic anomaly detection. 
	For the visual context representation, considerable research \cite{10.1145/3503161.3549201, Liang_2023_ICCV, Zhou_2022_CVPR} has demonstrated that CLIP possesses a robust foundation of vision-language prior knowledge. Leveraging this acquired semantic knowledge for anomaly detection in driving videos facilitates the perception and comprehension of driving behavior. Therefore, we advocate applying the pretrained visual encoder of CLIP to extract visual representations from driving video clips of two consecutive frames. 
	After obtaining the frame representations, we employ \textit{Average Pooling} along the temporal dimension as in previous works \cite{wang2021actionclip, LUO2022293, Rasheed_2023_CVPR} to aggregate these representations to characterize the visual context of the video clip. 
	For the text representation, we first describe normal and abnormal traffic events as text prompts (\textit{i.e.}, $a_1$ and $a_2$ in Table \ref{prompts}), and then apply the pretrained textual encoder in CLIP to extract text representations.
	
	Intuitively, after extracting the visual and textual representations of driving video clips, we can directly leverage contrastive learning to align them for traffic anomaly detection. However, in our task, solely modeling the visual representation from visual context is insufficient to capture the dynamic changes in the driving scene. Therefore, we introduce temporal high-frequency modeling (THFM) to characterize the dynamic changes and provide a more comprehensive representation of the driving video clips. Additionally, to better perceive various types of traffic anomalies, we further propose an attentive anomaly focusing mechanism (AAFM) to adaptively focus on the visual context of interest in the driving scene, thereby facilitating the detection of traffic anomalies.
	In the following sections, we will introduce these two key modules in detail.
	\subsection{Temporal High-Frequency Modeling (THFM)}
\begin{figure}[h]
		\centering
		\includegraphics[width=0.95\linewidth]{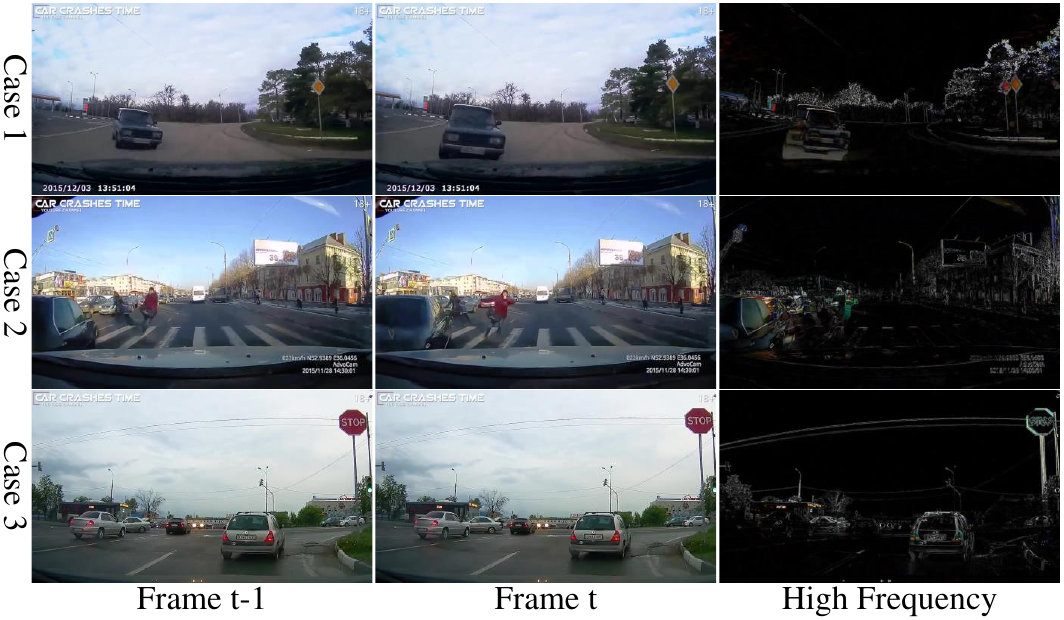}
		\caption{An illustration of the high frequency. We show 3 cases as examples. The first and second columns correspond to the original consecutive video frames, and the last column is the high-frequency component extracted along the temporal dimension.}
		\label{fig:diff}
	\end{figure}

	Video-text alignment diverges from image-text alignment by necessitating consideration of temporal characteristics. Numerous methods \cite{wang2021actionclip, LUO2022293, Rasheed_2023_CVPR} have effectively employed CLIP in addressing downstream tasks within the video domain. The modeling strategies adopted in these approaches for the temporal domain encompass various techniques such as \textit{Average Pooling}, \textit{Conv1D}, \textit{LSTM}, and \textit{Transformer}. These strategies primarily emphasize aggregating visual context from distinct video frames along the temporal dimension. Nevertheless, for the anomaly detection task in driving videos, we contend that not only the visual context but also the temporal dynamic changes in the driving scene hold significant importance in modeling driving behavior. For instance, a collision or loss of vehicle control often induces substantial changes in the driving scene within a brief timeframe. 
	Therefore, in our work, we propose to model the visual representation of driving videos in two aspects, \textit{i.e.}, the visual context of video frames in the spatial domain and the dynamic changes of driving scenes in the temporal domain.
	\begin{figure*}[!t]
		\centering
		\includegraphics[width=0.95\linewidth]{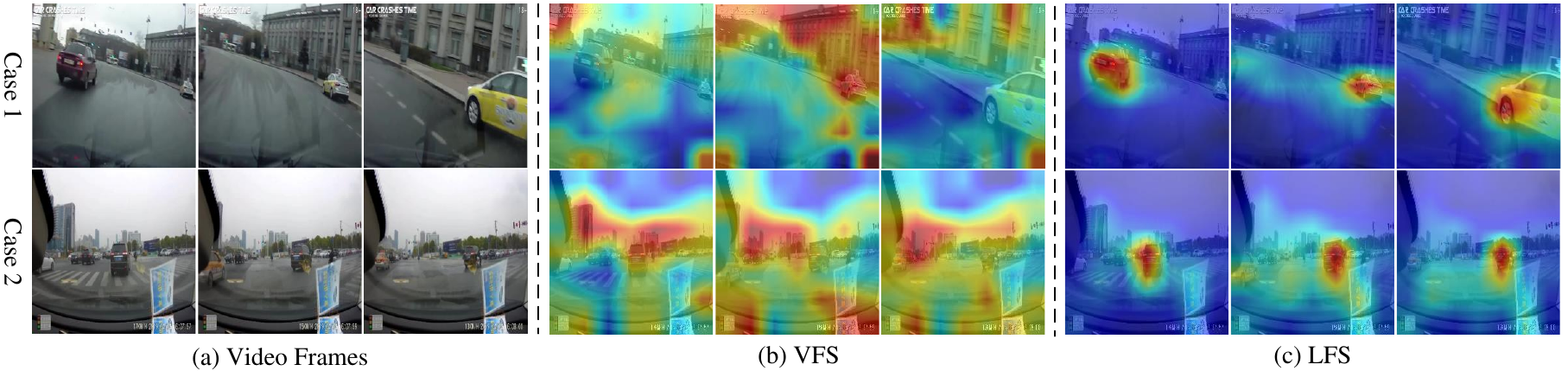}
		\caption{An illustration of the AAFM. The original video frames are displayed in column (a). In column (b), we visualize the attention of the visual representation to the deep features of a video clip under the visually focused strategy (VFS). In column (c), we visualize the attention of the soft text representation to the deep features of a video clip under the linguistically focused strategy (LFS). We present two types of traffic anomaly scenarios. Specifically, case 1 illustrates an instance where the ego-vehicle experiences loss of control while executing a turn. In case 2, the driving vehicle observes a collision between the car turning ahead and the motorcycle traveling straight on the right.}
		\label{fig:aafm}
	\end{figure*}
	Considering the fact that the high frequency of the driving video in the temporal domain reflects the dynamic changes of the driving scene. To clarify, we present several cases in Fig. \ref{fig:diff} for illustration.
	Based on the above observations, we introduce the Temporal High Frequency Modeling (THFM) to enhance the visual representation of the driving video within the temporal-spatial domain.
	
	Our fundamental idea involves utilizing the high frequency presented in the temporal domain of the driving video to characterize dynamic changes.
	Specifically, we first extract the high frequency of the driving video clip in the temporal dimension, which is formulated as:
	\begin{equation}
	I_{n}^{hp} = HP (frame_n^{t}, frame_n^{t-1}),
	\end{equation}
	where $HP(\cdot)$ is the difference operation to extract high frequency $I_{n}^{hp}$ along the temporal dimension from two consecutive frames $t-1$ and $t$ of the $n$-th driving video clip. Further, we encode $I_{n}^{hp}$ to the high-frequency representation by
	\begin{equation}
	H_{n}^{t} = F_{hf} (I_{n}^{hp}; \xi _{hf}),
	\end{equation}
	where $F_{hf}(\cdot)$ represents the high-frequency encoder, sharing the same architecture as the visual encoder (\textit{i.e.}, ResNet50 unless specified otherwise). The resultant high-frequency representation is denoted as $H_{n}^{t}$. Finally, to obtain the visual representation of the driving video clip in the spatio-temporal domain, we fuse the spatial visual context representation with the temporal high-frequency representation $H_{n}^{t}$, which is expressed as follows:
	\begin{equation}
	\begin{aligned}
	I_{n}^{t}, I_{n}^{t-1} &= F_{ve} (frame_n^{t}, frame_n^{t-1}; \xi _{ve}), \\
	V_n^{t} &= AvgPool (I_{n}^{t}, I_{n}^{t-1}), \\
	F_n &= V_n^{t} + H_{n}^{t},
	\end{aligned}
	\end{equation}
	where $F_{ve}$ is the visual encoder with frozen pre-trained parameter $\xi _{ve}$, $I_{n}^{t}$ and $I_{n}^{t-1}$ represent visual representations of frame $t$ and $t-1$, respectively, and $V_n^{t}$ denotes the spatial visual context representation after \textit{Average Pooling}.Here, $F_n \in \mathbb{R}^{1 \times C}$ is the fused visual representation, where $C$ denotes the feature dimension. The fused visual representation $F_n$ not only models the visual context of driving video clips, but also characterizes the dynamic changes in the temporal dimension, which is beneficial for perception and understanding driving behaviors.
	\subsection{Attentive Anomaly Focusing Mechanism}
	Different types of traffic anomalies tend to exhibit distinct characteristics. For instance, anomalies involving the ego vehicle are often accompanied by global jitter from the dashboard camera, whereas anomalies involving non-ego vehicles typically cause anomalies in local regions of the driving scene. Blindly encoding the entire driving scene may reduce the discriminability of driving events and impede the ability to detect various types of traffic anomalies. Therefore, adaptively focusing on the visual context of interest is critical to perceiving different types of traffic anomalies.
	
	In our work, we propose an attentive anomaly focusing mechanism (AAFM). 
	The fundamental idea is to decouple the visual context visually and linguistically, to guide the model to adaptively focus on the visual content of interest. Specifically, we carefully design two focusing strategies: the \textit{visually focused strategy (VFS)} and the \textit{linguistically focused strategy (LFS)}. The former utilizes visual representations with global context to concentrate on the most semantically relevant visual context, while the latter adaptively focuses on visual contexts that are most relevant to text prompts through the guidance of language.
	\subsubsection{Visually Focused strategy (VFS)} \label{vfr}
	In fact, the spatial visual representation inherently captures the global context. Utilizing the attention of visual representation towards the deep features of various regions in the driving scene enables a focus on the most semantically relevant visual content.
	Specifically, as shown in Fig. \ref{fig:framework}, we focus on and weight the deep features of interest by using cross-attention (CA) on the spatial visual context representation $V_n^{t}$ and deep features of the video clip, which can be written as:
	\begin{equation}
	VFR_n = softmax \left( \frac{Q\left( V_{n}^{t} \right) \cdot K^{\top}\left( P \right)}{c} \right) \cdot V\left( P \right),
	\end{equation}
	where $Q$, $K$ and $V$ are linear transformation, $P \in \mathbb{R}^{h*w \times C}$ is the deep feature map of the video clip, $(h, w)$ represents the size of the feature map, and $c$ is the scaling factor which refers to the rooted square of feature dimension. Note that, for transformer-based visual encoders, $V_{n}^{t}$ is represented by the class token, and $P$ is represented by the patch tokens. $VFR_n \in \mathbb{R}^{1 \times C} $ denotes the visually focused representation of the $n$-th video clip. Since the spatial visual representation encodes global context, focusing on its most relevant visual content helps guide the model to perceive the semantics of the driving scene. As shown in Fig. \ref{fig:aafm} (b), our VFS can adaptively focus on the crucial scene semantics in the driving scene. Such attention helps to detect traffic anomalies involving the ego-vehicle, especially the loss of control of the ego vehicle (case 1 in Fig. \ref{fig:aafm}).
	\subsubsection{linguistically focused strategy (LFS)}
	
	Intuitively, the fine-grained text prompts clearly define the subjects, objects, and traffic types involved in the traffic events. In contrast to general text prompts (as listed in $a_1$ and $a_2$ in Table \ref{prompts}), utilizing fine-grained text prompts helps guide the model to focus on relevant visual contexts, thereby improving the comprehension of various traffic anomalies.
		Therefore, to facilitate the model’s adaptive perception of relevant visual context, we further design a linguistically focused strategy. The core idea is to utilize the carefully designed fine-grained text prompts (as listed in $b_1$ to $b_4$ in Table \ref{prompts}) to guide the model to adaptively focus on the visual context of interest, thereby enhancing the understanding of traffic anomalies.
	
	Specifically, first, we categorize traffic events into four groups based on their types. Second, we further categorize each type of traffic event according to the different subjects (\textit{i.e.}, ego or non-ego vehicle) and objects (\textit{i.e.}, vehicle, pedestrian, or obstacle) involved. Finally, we define a total of 11 types of fine-grained text prompts, as summarized in Table \ref{prompts} from $b1$ to $b4$. Note that the DoTA dataset used in our experiments is annotated with 9 types of traffic anomalies, as shown in Table \ref{categories}, with each anomaly encompassing both ego-involved and non-ego traffic anomalies.
	With the defined fine-grained text prompts, we apply the textual encoder in CLIP to extract the fine-grained text representation as follows: 
	\begin{equation}
	T_m^{'} = F_{te} (t_m; \xi _{te}),
	\end{equation}
	where $F_{te}$ is the textual encoder with parameter $\xi _{te}$, $t_m$ ($m \in [1,11] \cap \mathbb{Z}$) denotes the $m$-th fine-grained text prompt, and $T_m^{'}$ represents the corresponding text representation. As we can see, the fine-grained text prompts describe the subjects and objects involved in a traffic event in a video frame, as well as the event type, which helps to focus on the visual regions in the driving scene where the traffic event occurred. Therefore, we further propose to leverage the similarity of the fine-grained text representation with each deep feature of the video clip to focus on the most relevant visual context of the text prompt. Note that in the driving scenario, we do not have direct access to realistic text prompt that match the driving video. To solve this problem, we leverage the similarity between the visual representation $F_n$ and fine-grained text representations to weight the text representations, and obtain the soft text representation as follows:
	\begin{equation}
	\begin{aligned}
	A_{n}^{m}=\frac{F_n\cdot (T_{m}^{'})^{\top}}{\lVert F_n \rVert \cdot \lVert T_m \rVert}, \\
	T_{soft} = \sum_{m=1}^{11}{A_{n}^{m}}T_{m}^{'},
	\end{aligned}
	\end{equation}
	where $A_n^{m}$ is the cosine similarity between the $n$-th visual representation $F_n$ and the $m$-th fine-grained text representation $T_{m}^{'} \in \mathbb{R}^{1 \times C}$. After obtaining the soft text representation $T_{soft} \in \mathbb{R}^{1 \times C}$, similar to Section \ref{vfr}, we can further focus on the most semantically relevant visual context of the text description based on the cross-attention (CA) on the soft text representation $T_{soft}$ and deep features $P$, which is denoted as:
	\begin{equation}
	LFR_n=softmax \left( \frac{Q\left( T_{soft} \right) \cdot K^{\top}\left( P \right)}{c} \right) \cdot V\left( P \right). 
	\end{equation}
	$LFR_n \in \mathbb{R}^{1 \times C} $ represents the linguistically focused representation of the $n$-th video clip, which focuses on the visual context that is most relevant to the soft text representation $T_{soft}$. Moreover, Fig. \ref{fig:aafm}(c) shows that our LFS can indeed adaptively concentrate on road participants potentially linked to anomalies. This capability is crucial for identifying local anomalies in driving scenarios arising from non-ego vehicles (case 2 in Fig. \ref{fig:aafm}).
	
	Finally, we enhance the visual representation $F_n$ of driving videos by fusing it with visually and linguistically focused representations. Formally, it can be expressed as:
	\begin{equation}
	F_{n}^{'}=F_{fusion} ([VFR_n, LFR_n, F_n]; \xi _f) ,
	\end{equation}
	where $F_{fusion}$ is the fusion layer composed of multi-layer perceptrons with parameter $\xi _f$. $F_{n}^{'}$ is an enhanced visual representation that not only adaptively focuses on the visual contexts of interest but also more comprehensively characterizes the driving video clip in the spatio-temporal domain. Moreover, such representations facilitate the alignment of visual representations with general text prompts, thus improving the detection of traffic anomalies.
	\begin{table}[!t]
		\renewcommand{\arraystretch}{1.2}
		\centering
		\caption{Summary of well-designed text prompts.}
		\begin{tabular}{c | c m{5.5 cm}}
			\hline
			\multirow{2}{*}{\parbox{1.5cm}{\centering {General \newline {Text Prompt}}}}& $a_1$: & \textit{``A traffic anomaly occurred in the scene."}\\
			& $a_2$: & \textit{``The traffic in this scenario is normal."}\\ \hline
			\multirow{4}{*}{\parbox{1.5cm}{\centering {Fine-grained \newline{Text Prompt}}}}& $b_1$: &  \textit{``The \{ego, non-ego\} vehicle collision with another \{vechile, pedestrian, obstacle\}."}\\
			& $b_2$: &\textit{``The \{ego, non-ego\} vehicle out-of-control and leaving the roadway."}\\
			& $b_3$: &\textit{``The \{ego, non-ego\} vehicle has an unknown accident."}\\ 
			& $b_4$: &\textit{``The vehicle is running normally on the road."}\\
			\hline
		\end{tabular}
		\label{prompts}
	\end{table}
	\subsection{Contrastive Learning Strategy and Inference Process}
	In this section, we introduce the contrastive learning strategy of the proposed TTHF framework for cross-modal learning and present how to perform traffic anomaly detection. 
	
	Suppose that, there are $N$ video clips in the batch, we denote:
	\begin{equation}
	\begin{aligned}
	F=\{F_1, F_2, ..., F_N \}, \\
	F^{'}=\{F_{1}^{'}, F_{2}^{'}, ..., F_{N}^{'}\}, 
	\end{aligned}
	\end{equation}
	where $F$ is the visual representation of $N$ video clips and $F^{'}$ represents the enhanced visual representation. For text prompts, we denote:
	\begin{equation}
	\begin{aligned}
	T=\{T_1, T_2, ..., T_{N}\}, \\
	T^{'}=\{T_1^{'}, T_2^{'}, ..., T_N^{'}\}, 
	\end{aligned}
	\end{equation}
	where $T$ means the matched general text representation of $N$ video clips and $T^{'}$ is the matched fine-grained text representation. Note that $T_n$ and $T^{'}_n$ denote the high-dimensional representations of one of the $D$ predefined text prompts. In our case, $D=2$ for general text prompts and $D=11$ for fine-grained text prompts. To better understand abstract concepts of traffic anomalies, we first perform contrastive learning to align visual representations $F$ with fine-grained text representations $T^{'}$. Formally, the objective loss along the visual axis can be expressed as:
	\begin{equation}
	\mathcal{L}_{vf}=\frac{1}{N}\sum_{i=1}^N{-\log \frac{\exp ( F_i\cdot (T_{i}^{'})^{\top}/\tau )}{\sum_{j\in \left[1,D \right]}{\exp ( F_i\cdot (T_{j}^{'})^{\top}/\tau)}}}.
	\end{equation}
	For the $j$-th trained text representation $T_j$, it may actually match more than one visual representation. Symmetrically, we can calculate the loss along the text axis by:
	\begin{equation}
	\mathcal{L}_{tf}=\frac{1}{D}\sum_{j=1}^{D}{-\log \frac{\sum_{k\in \left[ 1,N \right] ,T_k^{'}=T_j^{'}}{\exp \left( T_k^{'}\cdot F_{k}^{\top}/\tau \right)}}{\sum_{i\in \left[ 1,N \right]}{\exp \left( T_j^{'}\cdot F_{i}^{\top}/\tau \right)}}},
	\end{equation}
	where $\tau$ is a learned temperature parameter\cite{pmlr-v139-radford21a}. Similarly, we further apply contrastive learning to align the enhanced visual representations with the general text representations. The calculations along the visual and textual axis are as follows:
	\begin{equation}
	\begin{aligned}
	\mathcal{L}_{vg}=\frac{1}{N}\sum_{i=1}^N{-\log \frac{\exp ( F_i^{'}\cdot (T_{i})^{\top}/\tau )}{\sum_{j\in \left[1,D \right]}{\exp ( F_i^{'}\cdot (T_{j})^{\top}/\tau)}}}, \\
	\mathcal{L}_{tg}=\frac{1}{D}\sum_{j=1}^{D}{-\log \frac{\sum_{k\in \left[ 1,N \right] ,T_k=T_j}{\exp \left( T_k\cdot (F^{'}_{k})^{\top}/\tau \right)}}{\sum_{i\in \left[ 1,N \right]}{\exp \left( T_j\cdot (F^{'}_{i})^{\top}/\tau \right)}}}.
	\end{aligned}
	\end{equation}
	The overall loss then becomes:
	\begin{equation}
	\mathcal{L} = ((\mathcal{L}_{vf}+\mathcal{L}_{tf}) + (\mathcal{L}_{vg}+\mathcal{L}_{tg})) / 2.
	\end{equation}
	
	The inference procedure is similar to the training procedure. For the $i$-th testing driving video clip, our TTHF first extracts the visual representation $F_i$ and the enhanced visual representation $F_i^{'}$. For text prompts, the text encoder constructs $11$ fine-grained text representations $T^{'}=\{T_1^{'}, T_2^{'}, ..., T_{11}^{'}\}$ and $2$ general text representations $T=\{T_1, T_2\}$. We then compute the cosine similarity between $F_i$ and $T^{'}$ and between $F_i^{'}$ and $T$, respectively. Finally, we calculate the anomaly score for the $i$-th driving video clip as: 
	\begin{equation}
	Score_i = 1 - (S_f^{11} + S_g^{2}) / 2,
	\end{equation}
	where $S_f^{11}$ represents the cosine similarity after softmax between $F_i$ and $T_{11}^{'}$, and $S_g^{2}$ denotes the cosine similarity after softmax between $F_i^{'}$ and $T_2$. By taking the complement of the average over the prompts corresponding to normal traffic at different levels, we can obtain the final anomaly score $Score_i$.
	\section{Experiments and Discussions}  \label{sec:experiment}
	In this section, we evaluate the performance of our proposed method, which is performed on a platform with one NVIDIA 3090 GPU. All experiments were implemented using the PyTorch framework. Our source code and trained models will be publicly available upon acceptance. 
	
	\subsection{Implementation Details}
	In the experiments, we resize the driving video frames to $224 \times 224$ and take every two consecutive frames as the input video clip. Except where noted otherwise, in all experimental settings, we adopt ResNet-50 \cite{He_2016_CVPR} for the visual and high-frequency encoders and Text Transformer \cite{radford2019language} for the textual encoder. All of them are initialized with the parameters of CLIP's pre-trained model. Note that during the training phase, we freeze the pre-trained parameters of the visual encoder to prevent the model from overfitting to a specific dataset (\textit{e.g.}, DoTA) while enhancing the generalization of the visual representation. Besides, we optimize loss functions using the Adam algorithm with batch size $128$, learning rate 5e-6, weight decay 1e-4, and train the framework for $10$ epochs. During inference, we evaluate the traffic anomaly score by taking the complement of the similarity score of normal traffic prompts on both fine-grained and general text prompts. 
	\begin{table}[!t]
		\renewcommand{\arraystretch}{1.2}
		\centering  
		\caption{Traffic anomaly category in the DoTA dataset.}
		\label{categories}
		\resizebox{\linewidth}{!}{%
			\begin{tabular}{cc}
				\hline
				Label         & Anomaly Category                    \\ \hline
				ST & Collision with another vehicle that starts, stops, or is stationary \\
				AH & Collision with another vehicle moving ahead or waiting \\ 
				LA & Collision with another vehicle moving laterally in the same direction \\
				OC & Collision with another oncoming vehicle \\ 
				TC & Collision with another vehicle that turns into or crosses a road \\ 
				VP & Collision between vehicle and pedestrian \\
				VO & Collision with an obstacle in the roadway \\ 
				OO & Out-of-control and leaving the roadway to the left or right \\ 
				UK & Unknown \\ 
				\hline
		\end{tabular}}
	\end{table}
	\subsection{Dataset and Metrics}
	\begin{table}[!t]
		\renewcommand{\arraystretch}{1.1}
		\centering  
		\caption{The AUC $\uparrow$ (\%) of different approaches on the DoTA dataset.}
		\label{AUC}
		\begin{tabular}{c m{2cm} c c}
			\hline
			Methods                       & \centering Input        & Paradigm            & AUC (\%)  \\ \hline
			ConvAE \cite{Hasan_2016_CVPR} & \centering Gray         & Single-Stage             & 64.3 \\
			ConvAE \cite{Hasan_2016_CVPR} & \centering Flow         & Two-Stage              & 66.3 \\ 
			ConvLSTMAE \cite{10.1007}     & \centering Gray         & Single-Stage              & 53.8 \\
			ConvLSTMAE \cite{10.1007}     & \centering Flow         & Two-Stage             & 62.5 \\ 
			AnoPred \cite{Liu_2018_CVPR}  & \centering RGB          & Single-Stage              & 67.5 \\ 
			AnoPred \cite{Liu_2018_CVPR}  & \centering Mask RGB     & Two-Stage             & 64.8 \\
			FOL-STD \cite{8967556}        & \centering Box          & Two-Stage             & 66.7 \\ 
			FOL-STD \cite{8967556}        & \centering Box + Flow   & Two-Stage             & 69.1 \\ 
			FOL-STD \cite{8967556}        & \centering Box + Flow + Ego & Two-Stage         & 69.7 \\ 
			FOL-Ensemble \cite{9712446}   & \centering RGB + Box + Flow + Ego & Two-Stage   & 73.0 \\ 
			STFE \cite{9714213}           & \centering RGB + Box     & Two-Stage            & \underline{79.3} \\ \hline
			\bf{TTHF-Base}                    & \centering RGB           & Single-Stage          & 75.8 \\
			\bf{TTHF}                         & \centering RGB           & Single-Stage          & \bf{84.7} \\
			\hline
		\end{tabular}
	\end{table}
	\subsubsection{Dataset}
	For the sake of fairness, we evaluate our method on two challenging datasets, namely, DoTA \cite{9712446} and DADA-2000 \cite{9312486}, following prior works \cite{9712446, 9733965, 9714213}. 
	DoTA is the first traffic anomaly video dataset that provides detailed spatio-temporal annotations of anomalous objects for traffic anomaly detection in driving scenarios. The dataset contains 4677 dashcam video clips with a resolution of $1280 \times 720$ pixels, captured under various weather and lighting conditions. Each video is annotated with the start and end time of the anomaly and assigned to one of nine categories, which we summarize in Table \ref{categories}. The DADA-2000 dataset consists of 2000 dashcam videos with a resolution of $1584 \times 660$ pixels, each annotated with driver attention and one of 54 anomaly categories. In our experiments, we use the standard train-test split as used in \cite{9712446, 9312486} and other previous works.
	\begin{table*}[!t]
		\renewcommand{\arraystretch}{1.1}
		\centering 
		\caption{The AUC $\uparrow$ (\%) of different methods for each individual anomaly class on the DoTA dataset is presented. The $*$ indicates non-ego anomalies, while ego-involved anomalies are shown without $*$. 
			$N/A$ indicates that the AUC performance for the corresponding category is not available. We \textbf{bold} the best performance.} 
		\label{tab:my-table}
		\begin{tabular}{cccccccccccc} \hline 
			Methods & ST & AH & LA & OC & TC & VP & VO & OO & UK & AVG\\ \hline 
			AnoPred \cite{Liu_2018_CVPR} & 69.9 & 73.6 & 75.2 & 69.7 & 73.5 & 66.3 & N/A & N/A & N/A & 71.4\\ 
			AnoPred \cite{Liu_2018_CVPR}+Mask & 66.3 & 72.2 & 64.2 & 65.4 & 65.6 & 66.6 & N/A & N/A & N/A & 66.7\\
			FOL-STD \cite{8967556} & 67.3 & 77.4 & 71.1 & 68.6 & 69.2 & 65.1 & N/A & N/A & N/A & 69.7\\ 
			FOL-Ensemble \cite{9712446} & 73.3 & 81.2 & 74.0 & 73.4 & 75.1 & 70.1 & N/A & N/A & N/A & 74.5\\ 
			STFE \cite{9714213} & \underline{75.2} & \underline{84.5} & 72.1 & \underline{77.3} & 72.8 & {71.9} & N/A & N/A & N/A & 75.6 \\
			\bf{TTHF-Base} & 72.8 & 79.6 & \underline{83.7} & 76.4 & \underline{82.6} & \underline{72.3} & \underline{81.8} & \underline{80.4} & \bf{72.7} & \underline{78.0}\\
			\bf{TTHF} & \bf{86.7} & \bf{90.5} & \bf{89.7} & \bf{87.0} & \bf{89.5} & \bf{77.1} & \bf{87.6} & \bf{90.1} & \underline{70.9} & \bf{85.5} \\
			\hline 
			Methods & ST* & AH* & LA* & OC* & TC* & VP* & VO* & OO* & UK* & AVG\\ \hline 
			AnoPred \cite{Liu_2018_CVPR} & 70.9 & 62.6 & 60.1 & 65.6 & 65.4 & 64.9 & 64.2 & 57.8 & N/A & 63.9 \\
			AnoPred \cite{Liu_2018_CVPR}+Mask & 72.9 & 63.7 & 60.6 & 66.9 & 65.7 & 64.0 & 58.8 & 59.9 & N/A & 64.1 \\
			FOL-STD \cite{8967556} & {75.1} & 66.2 & 66.8 & 74.1 & 72.0 & 69.7 & 63.8 & 69.2 & N/A & 69.6 \\
			FOL-Ensemble \cite{9712446}&  \underline{77.5} & 69.8 & 68.1 & \underline{76.7} & 73.9 & \underline{71.2} & 65.2 & 69.6  & N/A &  71.5 \\
			STFE \cite{9714213} & \bf{80.6} & 65.6 & \underline{69.9} & 76.5 & \underline{74.2} & N/A & \underline{75.6} & \underline{70.5} & N/A & \underline{73.2} \\
			\bf{TTHF-Base} & 75.0 & \underline{71.5} & 67.2 & 72.5 & 70.6 & 64.3 & 69.9 & 68.3 & \underline{68.1} & 69.7 \\
			\bf{TTHF} & 74.9 & \bf{76.0} & \bf{76.4} & \bf{79.8} & \bf{81.5} & \bf{79.2} & \bf{79.0} & \bf{77.5} & \bf{68.9} & \bf{77.0} \\
			\hline
		\end{tabular} 
	\end{table*}
	\subsubsection{Metrics}
	Following prior works \cite{9714213, Gong_2019_ICCV, 9712446}, we use Area under ROC curve (AUC) metric to evaluate the performance of different TAD approaches. The AUC metric is calculated by computing the area under a standard frame-level receiver operating characteristic (ROC) curve, which plots the true positive rate (TPR) against the false positive rate (FPR). The larger AUC prefers better performance.
	\subsection{Competitors}
	To verify the superiority of the proposed framework, we compare with the following state-of-the-art TAD approaches: ConvAE \cite{Hasan_2016_CVPR}, ConvLSTMAE \cite{10.1007}, AnoPred \cite{Liu_2018_CVPR}, FOL-STD \cite{8967556}, FOL-Ensemble \cite{9712446}, DMMNet \cite{9052726}, SSC-TAD \cite{9733965} and STFE \cite{9714213}. 
	Among them, the ConvAE \cite{Hasan_2016_CVPR} and ConvLSTMAE \cite{10.1007} methods contain two variants. The variant utilizing the grayscale image as input belongs to the single-stage paradigm, while the variant using optical flow as input belongs to the two-stage paradigm. The AnoPred method \cite{Liu_2018_CVPR} also contains two variants. The variant employing the full video frame as input falls within the single-stage paradigm, whereas the variant utilizing pixels of foreground objects belongs to the two-stage paradigm. Besides, the DMMNet method \cite{9052726} follows the single-stage paradigm, while the methods FOL-STD \cite{8967556}, FOL-Ensemble \cite{9712446}, SSC-TAD \cite{9733965}, and STFE \cite{9714213} fall under the two-stage paradigm. Note that the experimental results for all these methods and their variants are obtained from the published papers \cite{9712446, 9733965, 9714213}. In addition, we consider a CLIP-like TAD framework, denoted as TTHF-Base, as our baseline approach. This baseline lacks temporal High-Frequency Modeling and the attention anomaly focusing mechanism and utilizes only general text prompts for alignment.
	\begin{table}[!t]
		\renewcommand{\arraystretch}{1.1}
		\centering  
		\caption{The AUC $\uparrow$ (\%) of different methods on the DADA-2000 dataset.}
		\label{AUC_dada}
		\begin{tabular}{ccccc}
			\hline
			Methods                       & Trained & Ego-Involved              & Non-Ego  & Both\\ \hline
			AnoPred \cite{Liu_2018_CVPR}  & \checkmark & 55.7                      & 56.9     & 56.1\\ 
			FOL-STD \cite{8967556}        & \checkmark & 71.3                      & 57.1     & 66.6\\
			DMMNet \cite{9052726}         & \checkmark& 73.0                       & 56.3     & {67.5}\\
			SSC-TAD \cite{9733965}        & \checkmark & 67.6                      & {58.7}     & 66.5\\
			\bf{TTHF-Base}               & $\times$   & \underline{78.7}          & \underline{59.4}     & \underline{68.3}\\
			\bf{TTHF}                    & $\times$   & \bf{80.9}                 & \bf{64.0}        & \bf{71.7} \\
			\hline
		\end{tabular}
	\end{table}
	\begin{table}[!t]
		\renewcommand{\arraystretch}{1.1}
		\centering  
		\caption{Ablation results of different components on DoTA dataset. Note that for fair comparison, in the experiments without THFM, we fine-tune the parameters of the visual encoder. Larger AUC prefers better performance.}
		\label{ablation}
		\begin{tabular}{c c c c c c}
			\hline
			Arch.    & Visual         &  Textual        & AAFM        & THFM         & AUC (\%) \\ \hline
			TTHF    & \checkmark     &  $\times$       & $\times$   & $\times$      & 61.0 \\ 
			TTHF    & \checkmark     & \checkmark      & $\times$   & $\times$      & 75.8 \\ 
			TTHF    & \checkmark     & \checkmark      & \checkmark & $\times$      & 76.8 \\ 
			TTHF    & \checkmark     & \checkmark      & \checkmark & \checkmark    & \bf{84.7} \\ 
			\hline
		\end{tabular}
	\end{table}
	\begin{table}[!t]
		\renewcommand{\arraystretch}{1.1}
		\centering  
		\caption{Ablation results on how AAFM contributes to traffic anomaly detection on the DoTA dataset. Larger AUC prefers better performance.}
		\label{aafm}
		\begin{tabular}{c c c c}
			\hline
			Arch.       & VFS        & LFS         & AUC (\%) \\ \hline
			TTHF       & $-$        & $-$         & 75.8 \\ 
			TTHF       & \checkmark & $\times$    & 76.3 \\ 
			TTHF       & $\times$   & \checkmark  & 76.5 \\ 
			TTHF       & \checkmark & \checkmark  & \bf{76.8} \\ 
			\hline
		\end{tabular}
	\end{table}
	\begin{table}[!t]
		\renewcommand{\arraystretch}{1.1}
		\centering  
		\caption{Ablation results of different backbones on DoTA dataset. Larger AUC prefers better performance.}
		\label{backbone}
		\begin{tabular}{c|c m{1.5cm} |c}
			\hline
			Arch.    & Visual    & \centering Textual               & AUC (\%) \\ \hline
			TTHF    & RN-50     & \centering Text-Transformer      &  84.7 \\ 
			TTHF    & RN-50x64     & \centering Text-Transformer      & 84.8 \\
			TTHF    & ViT-B-32     & \centering Text-Transformer      & 84.0 \\
			TTHF    & ViT-L-14     & \centering Text-Transformer      & \bf{85.0} \\ 
			\hline
		\end{tabular}
	\end{table}
	\begin{figure*}[!t]
		\centering
		\includegraphics[width=0.97\linewidth]{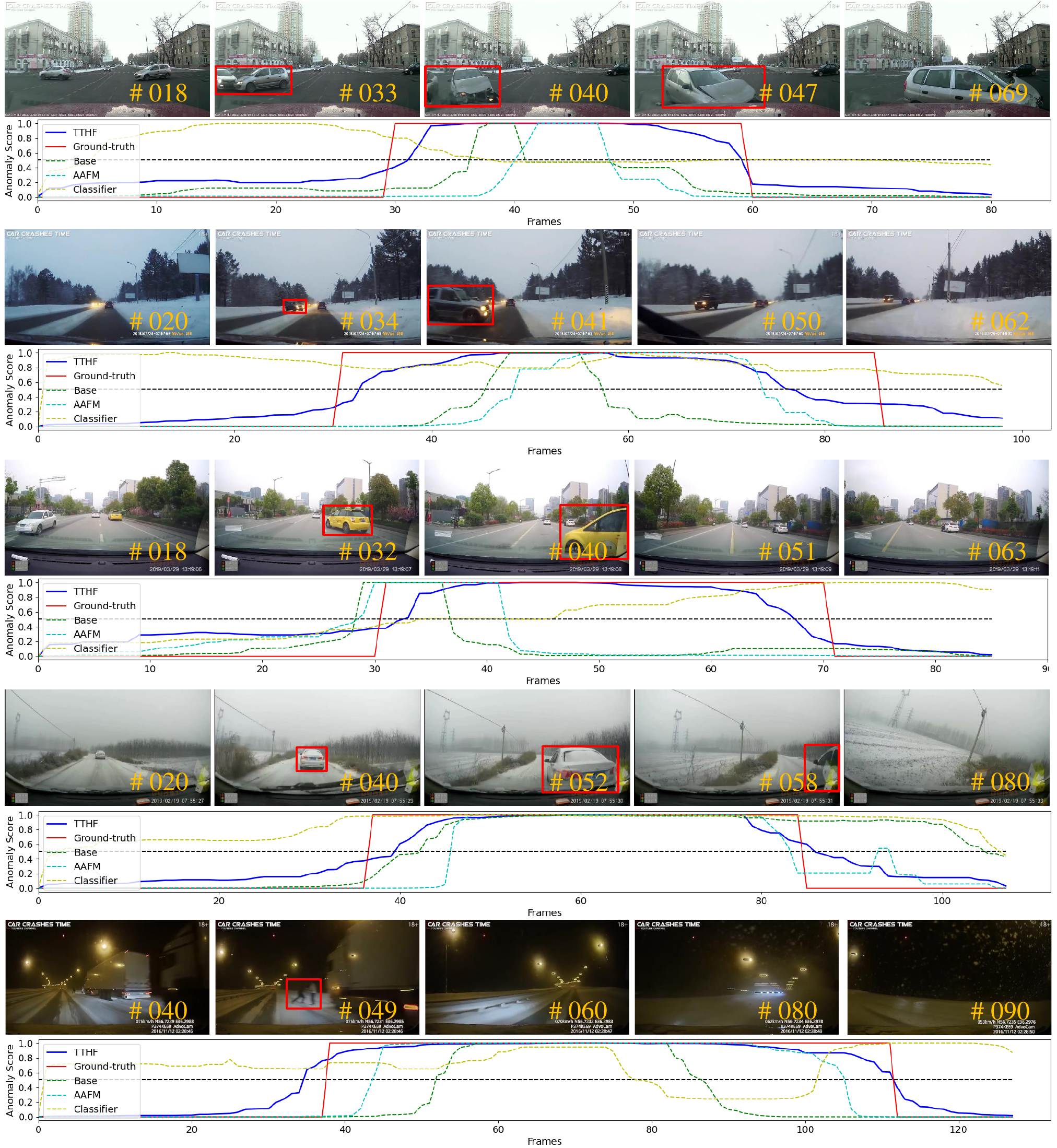}
		\caption{The visualization of anomaly score curves for traffic anomaly detection of different variants on the DoTA dataset. The first row of each case shows the extracted video frames of the driving video, where the red boxes mark the object involved in or causing the anomaly. The second rows show the anomaly score curves of different methods on the corresponding whole videos. For brevity, we label the TTHF-Base variant as Base and TTHF-Base with AAFM as AAFM, while Classifier denotes the classify-based TAD method. Better viewed in color.}
		\label{fig:vis}
	\end{figure*}
	\subsection{Quantitative Results}
	\subsubsection{Overall results} We conduct a comparative analysis of TTHF with a wide range of competitors and their variants in terms of AUC metric. Table \ref{AUC} presents the AUC performance of various competitors, along with labels indicating their respective variants (\textit{i.e.}, different inputs) and paradigms employed. Overall, our framework demonstrates the superior performance on the DoTA dataset in terms of AUC. Specifically, our method outperforms the previously two-stage paradigm-based leading TAD method, STFE \cite{9714213}, by +5.4\% AUC. Although in previous methods, the two-stage paradigm method employs a perception algorithm in the first stage to mitigate the impact of dynamic background resulting from the ego-vehicle movement, and generally outperforms single-stage TAD methods \cite{Hasan_2016_CVPR, 10.1007,Liu_2018_CVPR}, such approaches are susceptible to the performance of the perception algorithm in the first stage, potentially leading to error propagation. In contrast, our proposed single-stage TAD method explicitly characterizes dynamic changes by modeling high frequency in the temporal domain, achieving a significant performance improvement over all previous methods and establishing a new state-of-the-art in traffic anomaly detection. Note that our baseline method outperforms all previous single-stage paradigm-based methods by at least +8.3\% AUC. This is mainly attributed to our introduction of text prompts and the alignment of driving videos with text representations in a high-dimensional space, which facilitates the detection of traffic anomalies.
	\begin{figure*}[ht]
		\centering
		\includegraphics[width=0.95\linewidth]{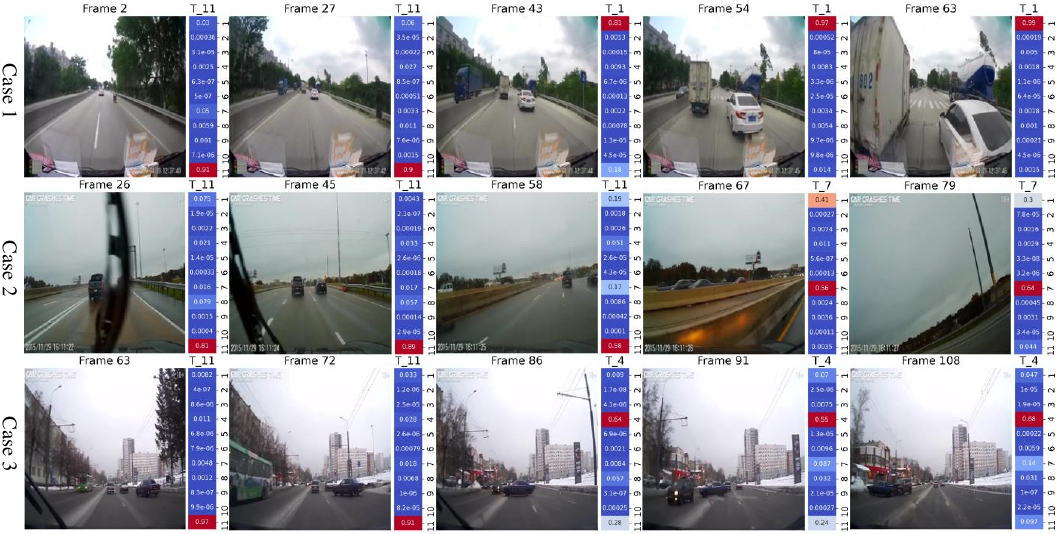}
		\caption{Visualization of the weights used for computing soft text representations. We present three illustrative cases, each involving video frames captured at different times. These frames are accompanied by the corresponding weight values used in the computation of soft text representations. Notably, we employ a blue-to-red color scale, where increasing redness signifies higher weights. Additionally, we label the ground-truth fine-grained text representations (denoted as T\_i) associated with specific frames. Among them, T\_1 corresponds to the text \textit{``The ego vehicle collision with another vehicle"} (as described in Table \ref{prompts}), T\_4 corresponds to the text \textit{``The non-ego vehicle collision with another vehicle"}, T\_7 corresponds to the text \textit{``The ego vehicle out-of-control and leaving the roadway"}, and T\_11 corresponds to the text \textit{``The vehicle is running normally on the road"}.}
		\label{fig:soft}
	\end{figure*}
	\subsubsection{Per-class results}
	To investigate the ability of our proposed method to detect traffic anomalies in different categories, we compared the detection performance of different methods for ego-involved and non-ego traffic anomalies. Based on the nine traffic anomalies divided by the DoTA dataset, detailed in Table \ref{categories}, we summarize the AUC performance of the different methods as well as the average AUC in Table \ref{tab:my-table}. Our method achieves significant improvements in all categories of traffic anomalies except ST*, and in particular, achieves an average AUC of at least +9.9\% on egos involving traffic anomalies. This further validates our idea that characterizing dynamic changes in driving scenarios is important for traffic anomaly detection. Simultaneously, it also demonstrates the effectiveness of our proposed approach to model the temporal high frequency of driving videos to characterize the dynamic changes of driving scenes.
	\begin{figure*}[ht]
		\centering
		\includegraphics[width=0.97\linewidth]{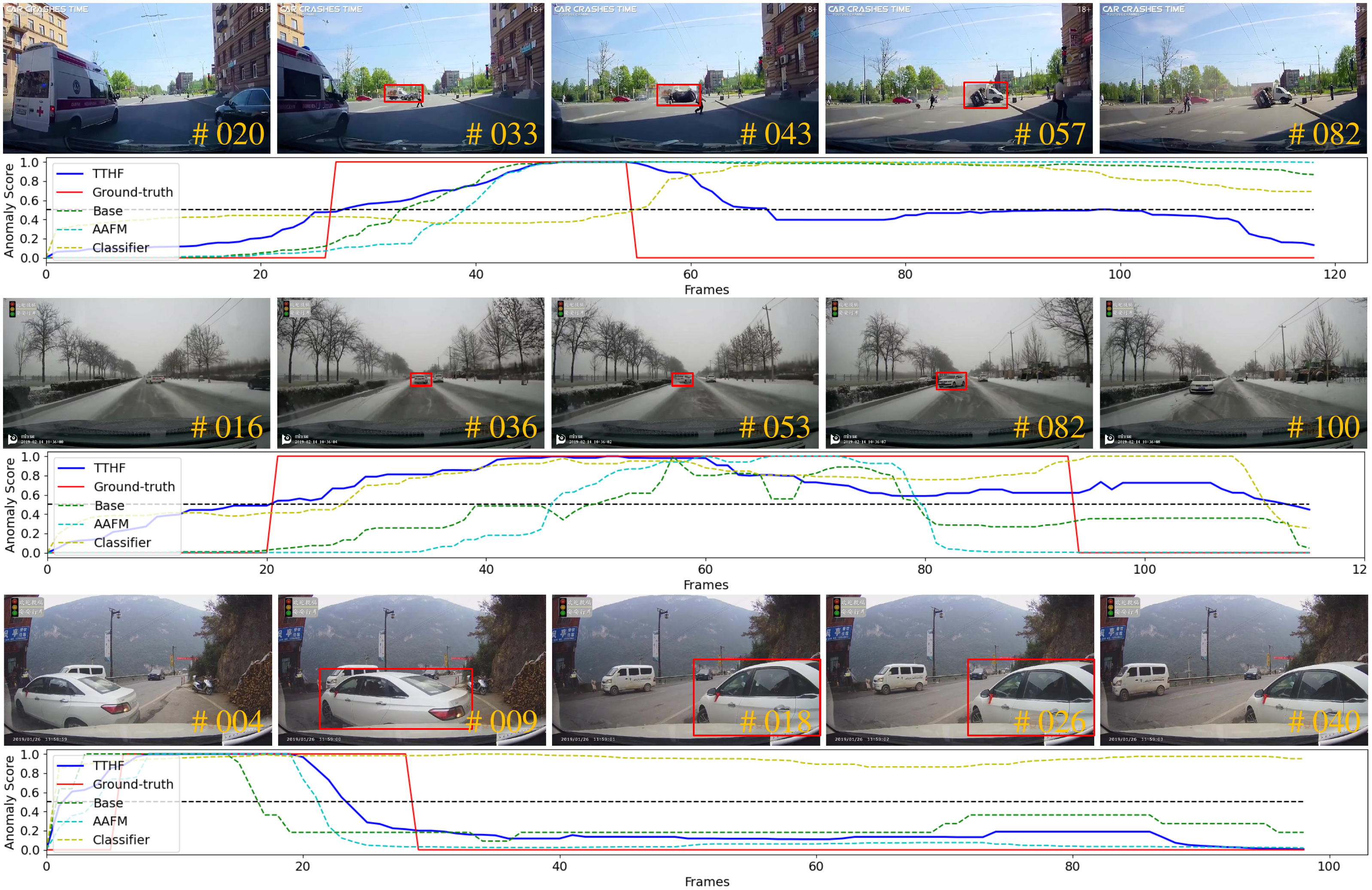}
		\caption{Visualization of some bad cases of the proposed TTHF. The first row of each case shows the extracted video frames of the driving video, where the red boxes mark the objects involved in the anomaly. The second rows show the anomaly score curves of different methods on the corresponding whole videos. Better viewed in color.}
		\label{fig:dis}
	\end{figure*} 
	\subsubsection{Generalization performance}
	To explore the generalization performance of our method for unseen types of traffic anomalies, we perform a generalization experiment on the DADA-2000 dataset. Specifically, we compare the AUC performance of our TTHF and TTHF-Base without any fine tuning on the DADA-2000 dataset with previous trained models, summarized in Table \ref{AUC_dada}. As we can see, 
	our proposed TTHF-base and TTHF methods outperform previously trained TAD methods, bringing at least +0.8\% and +4.2\% improvement in AUC respectively, indicating the strong generalization performance of the proposed approach.
	This is mainly attributed to our introduction of a text-driven video-text alignment strategy for traffic anomaly detection from a new perspective, as well as the proposed attentive anomaly focusing mechanism and temporal high-frequency modeling for traffic anomaly detection.
	\subsection{Qualitative Results}
	In this subsection, we visualize some examples to further illustrate the detection capability of our TTHF across various types of traffic anomalies and the feasibility of soft text representation in our framework.
	
	\subsubsection{Visualization of various types of traffic anomalies} As presented in Fig. \ref{fig:vis}, we show five representative traffic anomalies from top to bottom as examples: a) The other vehicle collides with another vehicle that turns into or crosses a road. b) The ego-vehicle collides with another oncoming vehicle. c) The ego-vehicle collides with another vehicle moving laterally in the same direction. d) The ego-vehicle collides with another vehicle waiting. e) The ego-vehicle is out-of-control and leaving the roadway to the left. From the above visualization results of different types of traffic anomalies, we can summarize as follows.
	Overall, our TTHF exhibits superior detection performance on various types of traffic anomalies. Secondly, while the most intuitive classify-based approach (It has the same network architecture as the visual encoder of TTHF, but directly classifies the visual representation, denoted as Classifier in  Fig. \ref{fig:vis}) also follows a single-stage paradigm, our proposed text-driven TAD approach offers a more comprehensive representation in high-dimensional space than orthogonal one-hot vectors. Consequently, both our proposed TTHF and its variants outperform the Classifier. Third, 
	incorporating AAFM allows our method to better perceive different types of traffic anomalies,
	as evident in Fig. \ref{fig:vis} when comparing the Base and AAFM variants across various traffic anomalies. Finally, capturing dynamic changes in driving scenarios significantly enhances traffic anomaly detection. This highlights the effectiveness of our approach in characterizing dynamic changes in driving scenarios by modeling high frequency in the temporal domain.
	\subsubsection{Visualization of the weights used for soft text representation} We further investigate the feasibility of soft text representations. Specifically, as shown in Fig. \ref{fig:soft}, we use three cases from the test set as examples. For video frames captured at different moments in driving videos, we visualize the weights employed to compute the soft text representation and compare it with the real fine-grained text representation. From the visualization results, we observe that the text representation associated with the maximum weight (indicated by the darkest red) consistently aligns with the real fine-grained text representation. The above results indicate that the way we calculate the soft text representation is effective and can well reflect the real anomaly category.
	\subsection{Ablation Investigation}
	In this subsection, we conduct ablation studies by analyzing how different components of TTHF contribute to traffic anomaly detection on DoTA dataset.
	\subsubsection{Variants of our architecture}
	We first evaluate the effectiveness of different components in our TTHF framework including the visual encoder, the textual encoder, the attentive anomaly focusing mechanism (AAFM), and the temporal high-frequency modeling (THFM). The ablation results are summarized in Table \ref{ablation}. Note that when only the visual encoder is applied, we add a linear classification head after the visual representation. This adaptation formulates the traffic anomaly detection task as a straightforward binary classification task. The results presented in Table \ref{ablation} demonstrate that introducing linguistic modalities and aligning visual-text in high-dimensional space greatly facilitates anomaly detection in driving videos compared to the classifier, achieving an AUC improvement of +14.8\%. Based on this, 
	the designed AAFM helps guide the model to adaptively focus on the visual context of interest and thus enhance the perception ability of various types of traffic anomalies. 
	Lastly, the incorporation of the modeling of temporal high frequency to capture dynamic background during driving significantly improves traffic anomaly detection, resulting in an AUC improvement of +7.9\%.
	\subsubsection{Analysis of the AAFM}
	To investigate how the proposed attentive anomaly focusing mechanism (AAFM) contributes to traffic anomaly detection, we perform ablation on each component in the AAFM. The ablation results are presented in Table \ref{aafm}. 
	We can conclude that both the Visually Focused Strategy (VFS) and the Linguistically Focused Strategy (LFS) explicitly guide the model to pay attention to the visual context most relevant to the representations of visual and linguistic modalities, respectively. 
	This enhances the ability to perceive traffic anomalies with different characteristics, thereby improving traffic anomaly detection in driving videos. Our AAFM achieves the best detection performance when both VFS and LFS are applied.
	\subsubsection{Network Architecture}
	Different network architectures of visual encoder may exhibit different representation capabilities. We now evaluate the performance of traffic anomaly detection when ResNet50 \cite{He_2016_CVPR}, ResNet50x64 \cite{pmlr-v139-radford21a}, ViT-B-32 \cite{dosovitskiy2020image} and ViT-L-14 \cite{dosovitskiy2020image} are used. Specifically, the results of these visual encoders can be found in Table \ref{backbone}, respectively. As can be noticed, for the task of traffic anomaly detection in driving videos, we observe that the ResNet-based network achieves comparable performance to the Transformer-based network. The larger model sizes perform slightly better, with ViT-L-14 achieving an AUC performance of 85.0\%. Therefore, considering both computing resources and performance gains, we ultimately chose ResNet50 as an example as our visual encoder in all other experiments.
	\subsection{Disscusion}
	In this subsection, we discuss the limitations of the proposed framework. We experimentally found that the detection accuracy of our proposed method needs improvement for two specific cases: 1) long-distance observation of traffic anomalies; and 2) subtle traffic anomalies involving other vehicles when the ego-vehicle is stationary. Fig. \ref{fig:dis} shows several cases where the accuracy of our method needs to be further improved. In the first scenario, the other vehicle at a distance collide with a turning or crossing vehicle. The second scenario depicts a distant vehicle losing control and veering to the left side of the road. The third scenario involves a slowly retreating vehicle experiencing friction with other stationary vehicles. By analyzing the anomaly score curve in Fig. \ref{fig:dis}, we can conclude that our method faces challenges primarily due to the traffic anomalies occurring in these scenarios involve non-ego vehicles and cause minor anomaly areas. These anomalies include small local anomalies that are caused when non-ego vehicles are abnormal at a distance, and slow and slight traffic anomalies that are observed for other vehicles when the ego-vehicle is at rest. These slight traffic anomalies may not be well focused on the corresponding abnormal regions by modeling the dynamic changes of the driving scene as well as using text guidance. This also explains that the ability of our method in detecting non-ego involved traffic anomalies is not as good as in detecting ego-involved traffic anomalies, especially ST* in Table \ref{tab:my-table}. Despite the significant improvement of our approach over previous TAD methods, addressing these more challenging traffic anomalies undoubtedly requires a greater effort from the community.
	\section{Conclusion}  \label{sec:conclusion}
	This paper have proposed an accurate single-stage TAD framework. For the first time, this framework introduces visual-text alignment to address the traffic anomaly detection task for driving videos. Notably, we verified that modeling the high frequency of driving videos in the temporal domain helps to characterize the dynamic changes of the driving scene and enhance the visual representation, thereby greatly facilitating the detection of traffic anomalies. In addition, 
	the experimental results demonstrated that the proposed attentive anomaly focusing mechanism is indeed effective in guiding the model to adaptively focus on the visual content of interest, thereby enhancing the ability to perceive different types of traffic anomalies.
	Although extensive experiments have demonstrated that the proposed TTHF substantially outperforms state-of-the-art competitors, more effort is required to accurately detect the more challenging slight traffic anomalies.
	
	\ifCLASSOPTIONcaptionsoff
	\newpage
	\fi
	
	\bibliography{TTHF.bib}
	\bibliographystyle{IEEEtran}
	\begin{IEEEbiography}
		[{\includegraphics[width=1in,height=1.25in,clip,keepaspectratio]{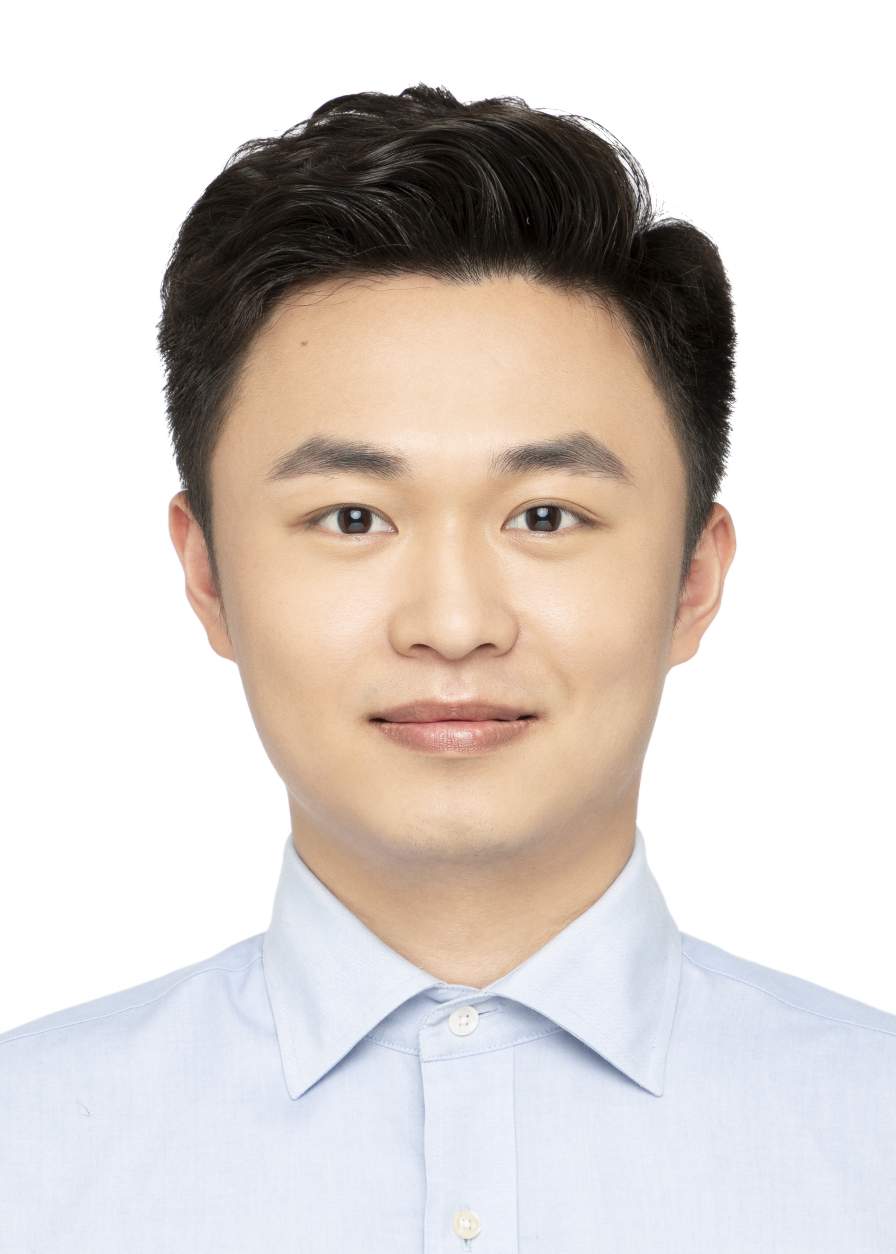}}]{Rongqin Liang} (Student Member, IEEE) received the B.Eng. degree in communication engineering from Wuyi University, Guangdong, China, in 2018 and M.S. degree in Information and Communication Engineering from Shenzhen University, Shenzhen, China, in 2021.
		He is currently a Ph.D. candidate at the College of Electronics and Information Engineering from Shenzhen University. His current research interests include trajectory prediction, anomaly detection, computer vision and deep learning.
	\end{IEEEbiography}
	\begin{IEEEbiography}
		[{\includegraphics[width=1in,height=1.25in,clip,keepaspectratio]{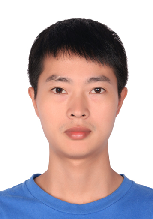}}]{Yuanman Li} (Senior Member, IEEE) received the B.Eng. degree in software engineering from Chongqing University, Chongqing, China, in 2012, and the Ph.D. degree in computer science from University of Macau, Macau, 2018. From 2018 to 2019, he was a Post-doctoral Fellow with the State Key Laboratory of Internet of Things for Smart City, University of Macau. He is currently an Assistant Professor with the College of Electronics and Information Engineering, Shenzhen University, Shenzhen, China. His current research interests include multimedia security and forensics, data representation, computer vision and machine learning.
	\end{IEEEbiography}
	\begin{IEEEbiography}
		[{\includegraphics[width=1in,height=1.25in,clip,keepaspectratio]{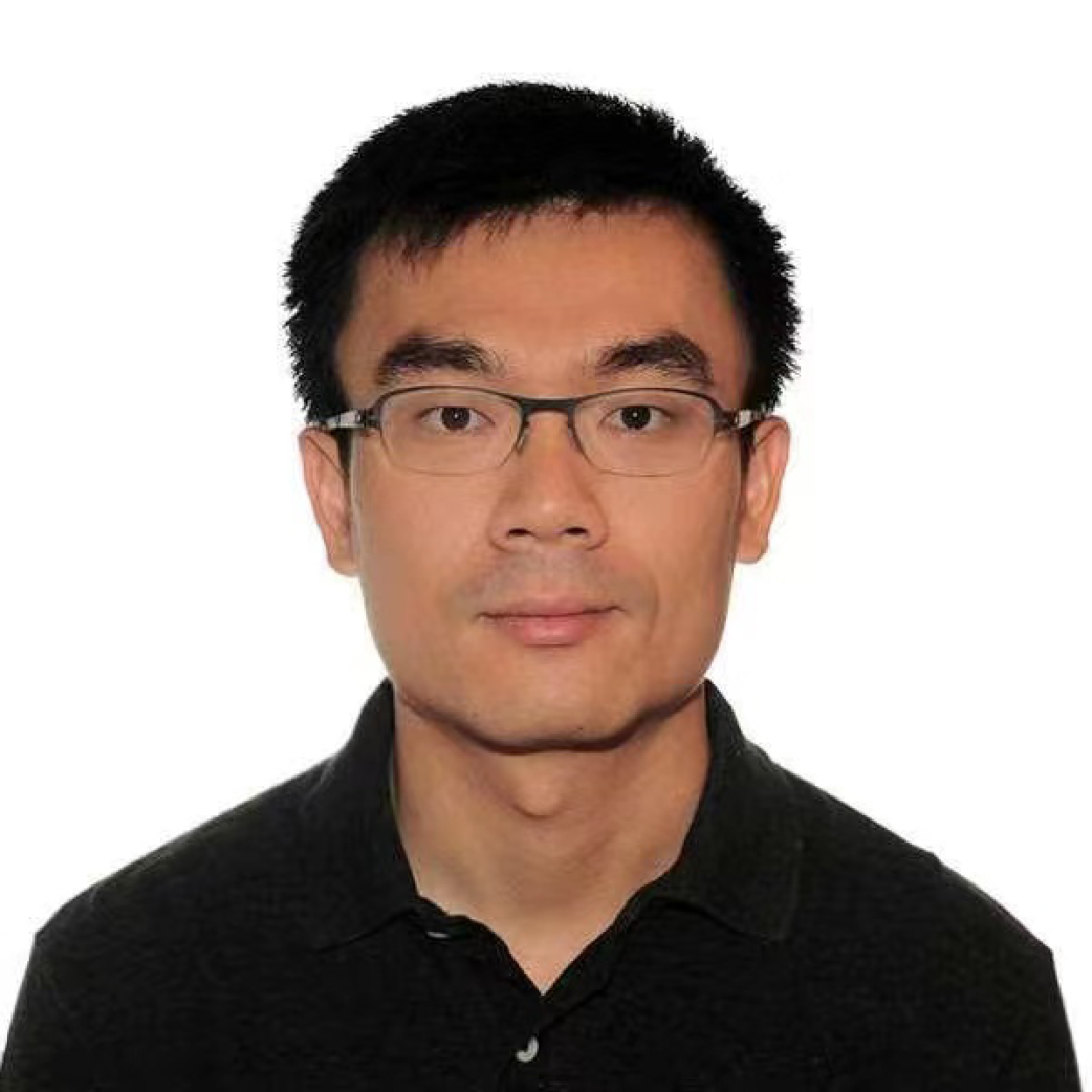}}]{Jiantao Zhou} (Senior Member, IEEE) received the B.Eng. degree from the Department of Electronic Engineering, Dalian University of Technology, in 2002, the M.Phil. degree from the Department of Radio Engineering, Southeast University, in 2005, and the Ph.D. degree from the Department of Electronic and Computer Engineering, Hong Kong University of Science and Technology, in 2009. He held various research positions with University of Illinois at Urbana-Champaign, Hong Kong University of Science and Technology, and McMaster University. He is an Associate Professor with the Department of Computer and Information Science, Faculty of Science and Technology, University of Macau, and also the Interim Head of the newly established Centre for Artificial Intelligence and Robotics. His research interests include multimedia security and forensics, multimedia signal processing, artificial intelligence and big data. He holds four granted U.S. patents and two granted Chinese patents. He has co-authored two papers that received the Best Paper Award at the IEEE Pacific-Rim Conference on Multimedia in 2007 and the Best Student Paper Award at the IEEE International Conference on Multimedia and Expo in 2016. He is serving as the Associate Editors of the IEEE TRANSACTIONS on IMAGE PROCESSING and the IEEE TRANSACTIONS on MULTIMEDIA.
	\end{IEEEbiography}	
	\vspace{-15 mm} 
	\begin{IEEEbiography}
		[{\includegraphics[width=1in,height=1.25in,clip,keepaspectratio]{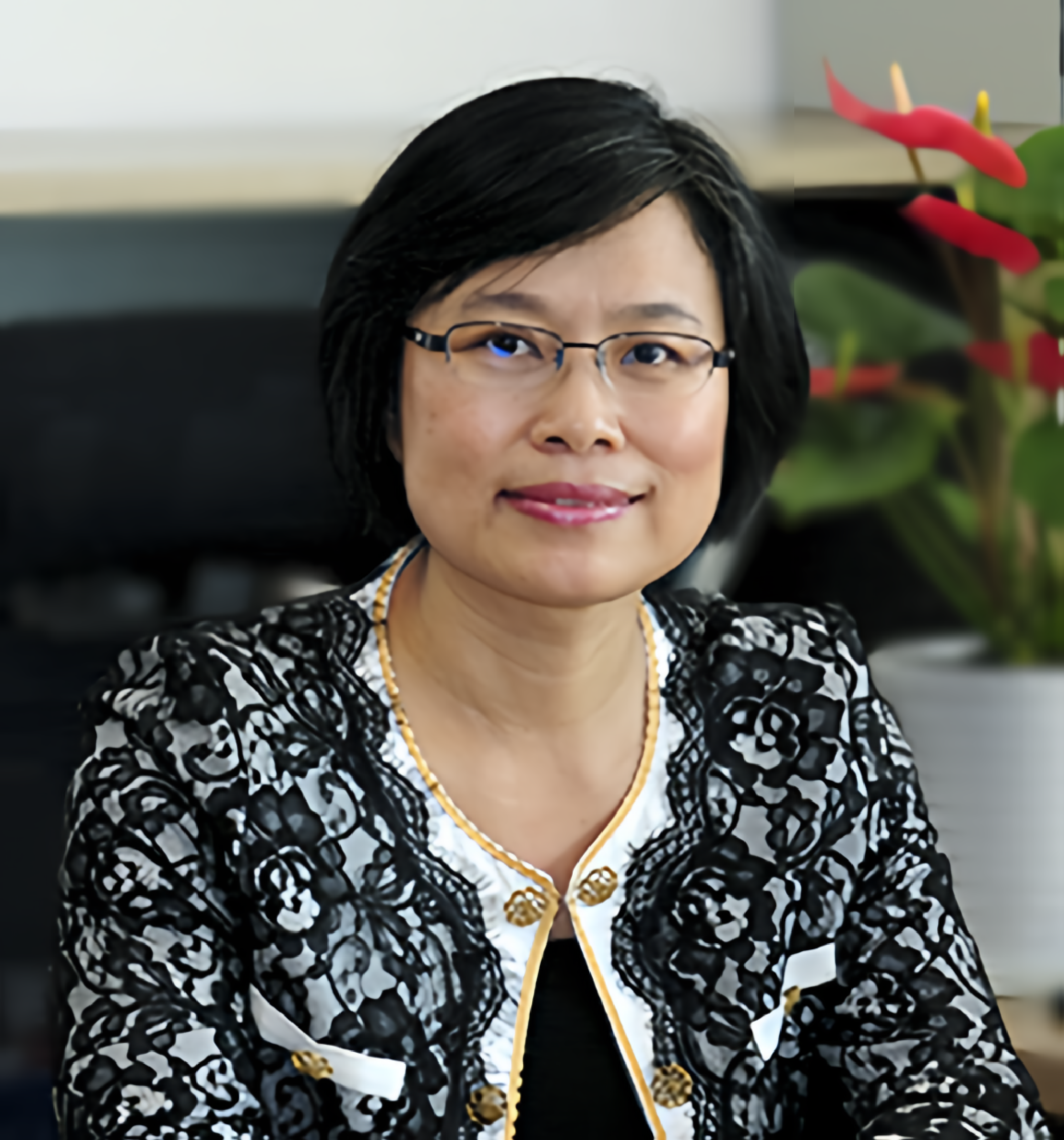}}]{Xia Li} (Member, IEEE) received her B.S. and M.S. in electronic engineering and SIP (signal and information processing) from Xidian University in 1989 and 1992 respectively. She was later conferred a Ph.D. in Department of information engineering by the Chinese University of Hong Kong in 1997. Currently, she is a member of the Guangdong Key Laboratory of Intelligent Information Processing. Her research interests include intelligent computing and its applications, image processing and pattern recognition.
	\end{IEEEbiography}
\end{document}